# Quasi-Direct Drive Actuation for a Lightweight Hip Exoskeleton with High Backdrivability and High Bandwidth

Shuangyue Yu*, Tzu-Hao Huang*, Xiaolong Yang, Chunhai Jiao, Jianfu Yang, Hang Hu, Sainan Zhang, Yue Chen, *Member, IEEE*, Jingang Yi, *Senior Member, IEEE*, Hao Su†, *Member, IEEE*

*Abstract*— High-performance actuators are crucial to enable mechanical versatility of lower-limb wearable robots, which are required to be lightweight, highly backdrivable, and with high bandwidth. State-of-the-art actuators, e.g., series elastic actuators (SEAs), have to compromise bandwidth to improve compliance (i.e., backdrivability). In this paper, we describe the design and human-robot interaction modeling of a portable hip exoskeleton based on our custom quasi-direct drive (QDD) actuation (i.e., a high torque density motor with low ratio gear). We also present a model-based performance benchmark comparison of representative actuators in terms of torque capability, control bandwidth, backdrivability, and force tracking accuracy. This paper aims to corroborate the underlying philosophy of "design for control", namely meticulous robot design can simplify control algorithms while ensuring high performance. Following this idea, we create a lightweight bilateral hip exoskeleton (overall mass is 3.4 kg) to reduce joint loadings during normal activities, including walking and squatting. Experimental results indicate that the exoskeleton is able to produce high nominal torque (17.5 Nm), high backdrivability (0.4 Nm backdrive torque), high bandwidth (62.4 Hz), and high control accuracy (1.09 Nm root mean square tracking error, i.e., 5.4% of the desired peak torque). Its controller is versatile to assist walking at different speeds (0.8-1.4 m/s) and squatting at 2 s cadence. This work demonstrates significant improvement in backdrivability and control bandwidth compared with state-of-the-art exoskeletons powered by the conventional actuation or SEA.

## I. INTRODUCTION

Safe and dynamic interaction with a human is of paramount importance for collaborative robots. State-of-the-art actuators (defined as motors with transmission), e.g., series elastic actuators (SEAs), have to compromise bandwidth to improve compliance (i.e., backdrivability) during interaction with a human. This limitation is particularly prominent for lower-limb wearable robots as it involves repeating high-force interaction with humans during locomotion with disturbance from changing gaits and varying terrain conditions [1-3]. Recent exoskeletons focus on advanced algorithms to improve control performance [4-9]. However, there is limited work in the actuation hardware design to provide the physical intelligence of robots, that is intrinsic to the system to ensure high performance without relying on complicated algorithms [10]. High-torque density electric motors represent a solution for physical intelligence to meet the multifaceted requirements of wearable robots ranging from lightweight, torque, bandwidth, backdrivability and force tracking accuracy.

Besides actuators for soft-material robots, there are three primary actuation methods for wearable robots, namely, conventional actuation, series elastic actuation (SEA), and quasi-direct drive actuation (also known as proprioceptive actuation [10, 11]). Conventional actuation uses high-speed and low-torque motors (typically brushless direct current motors, BLDC) with large ratio gears [12-15]. It can meet certain requirements such as assistive torque, angular velocity, and control bandwidth, but suffers from high mechanical impedance (i.e., resistive to human movements). Although control algorithms might be able to ameliorate the undesirable high impedance, it is infeasible to eliminate this side effect due to the inherent high inertia and high friction of actuators. Series elastic actuators (including parallel elastic actuators and other variable stiffness elastic actuators) overcome the backdrivability limitation [16-19] using the spring-type elastic elements. However, SEAs typically have to sacrifice the performance in control bandwidth, system complexity, system mass, and dimension, resulting in the limited practical benefits for wearable robots. Besides, the advantages of textile-based soft exosuits [4, 5, 20] (typically around 50:1 gear ratio) primarily originate from the wearable structure (textile) and cable transmission. Thus textile-based soft exosuits have similar high impedance limitations as SEA.

Quasi direct-drive (QDD) actuation composed of high torque motors and low gear ratio transmission represents a new solution to achieve versatility for wearable robots, namely high bandwidth and high backdrivability for a wide

This work is supported by the National Science Foundation grant IIS 1830613 and Grove School of Engineering, The City University of New York, City College. Any opinions, findings, and conclusions or recommendations expressed in this material are those of the author (s) and do not necessarily reflect the views of the funding organizations.

S. Yu, T. Huang, X. Yang, C. Jiao, J. Yang, S. Zhang, H. Hu, and H. Su are with Lab of Biomechatronics and Intelligent Robotics (BIRO), Department of Mechanical Engineering, The City University of New York, City College, NY, 10023, US (E-mail: hao.su@ccny.cuny.edu).

Y. Chen is with the Department of Mechanical Engineering, University of Arkansas, Fayetteville, AR, 72701, US

J. Yi is with the Department of Mechanical & Aerospace Engineering, Rutgers, The State University of New Jersey, Piscataway, NJ, 08854, US

*These authors contributed equally to this work. † corresponding author.

Digital Object Identifier (DOI): see top of this page.

variety of human activities. QDD actuation is recently studied for legged robots, but portable powered exoskeletons have disparate requirements in terms of mass, torque, and speeds. Wensing et al. [10] presented custom motors (interior rotor design, see section III-A) for cheetah robots, and the motor peak torque is 38.2 at 60 A electrical current. But the 1 Kg motor mass is too bulky and heavy for portable exoskeletons (overall weight of a bilateral exoskeleton with this motor is likely to be around 7 Kg including transmission, wearable structures, and battery). The mini cheetah robot [25] and commercial quasi-direct drive actuators for quadruped robots have high torque density because of exterior robot design (see Section III-A), but the nominal actuator torque is only 6-7 Nm, which is still not sufficient for lower limb exoskeleton applications. In the exoskeleton community, [21-24] presented tethered exoskeletons following the QDD paradigm and proved its feasibility. However, those systems are tethered and heavy, thus are not practical for real-world applications. Our recent work [25] presented a high-torque density electric motor (0.67 Nm nominal torque) with a low transmission ratio in a portable unilateral knee exoskeleton. The actuator can generate 6 Nm nominal torque and 16 Nm peak torque, and the device weighs 3.2 kg. Our hybrid soft exoskeleton for knee assistance [11] is tethered. Thus it is not suitable as personal mobility devices for community ambulation. Zhu et al. [26] proposed a knee exoskeleton with 2.69 kg unilateral mass using a quasi-direct drive actuation. The 1.15 kg actuator can generate a 10.6 Nm nominal torque and 20 Nm peak torque. The desire to further improve robot torque density and torque capability, to reduce the weight, and to investigate interaction modeling of this new actuation paradigm for portable exoskeleton devices motivates our work.

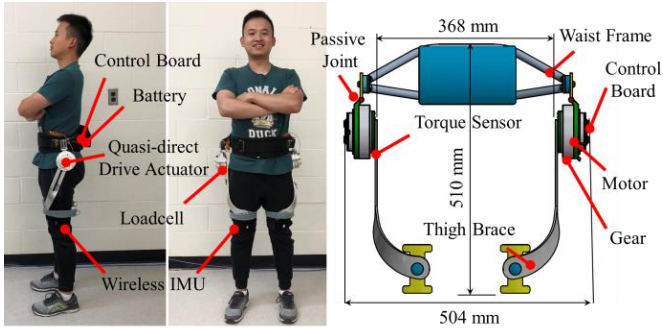

Fig. 1. The hip exoskeleton is versatile in terms of high bandwidth and high backdrivability thanks to its high torque density quasi-direct drive actuation. The mechanism of the hip exoskeleton is composed of a waist frame, two actuators, tow torque sensors, and two thigh braces. The passive joints allow the wearer's hip joint free abduction and adduction.

This paper presents a high-performance quasi-direct drive actuator using custom high torque density motor and low ratio gear transmission (8:1). The contributions of this paper include 1) a design method of quasi-direct drive actuation based exoskeleton that demonstrates mechanical versatility for being lightweight (3.4kg overall mass), highly-backdrivable (0.4 Nm backdrive torque) with high nominal torque (17.5 Nm) and high control bandwidth (62.4 Hz); 2) a unified human-robot interaction model that captures backdrivability and control bandwidth of three actuation paradigms (i.e., conventional actuator, series elastic actuator, quasi-direct drive actuator) with comprehensive benchmark results; 3) systematic characterization the quasi-direct drive actuator based hip exoskeleton in terms of output torque capability, control bandwidth, backdrivability, and torque control accuracy performance during walking and squatting assistance. *To the best of our knowledge, our hip exoskeleton has the best performance in terms of all key mechanical metrics, including nominal torque, backdrivability, bandwidth, and accuracy, as benchmarked with state-of-the-art hip exoskeletons [16-19] and hip exosuits [4, 5, 20]*. Though the focus of this paper is to reduce joint loadings of able-bodied populations for regular activities, our exoskeleton also has the potential to promote the independent living of people with lower-limb impairments.

## II. DESIGN REQUIREMENTS

The design of the hip exoskeleton needs to satisfy the kinematics and kinetics requirements. For walking, human hip joints have flexion/extension movements in the sagittal plane and abduction/adduction in the frontal plane. Therefore, the hip joint of the exoskeleton needs to accommodate those two degrees of freedom. For level-ground walking, the range of motion of a human hip joint is 32.2° flexion, 22.5° extension, 7.9° abduction, and adduction 6.4°. We design the robot with a larger range of motion than the standard requirements to handle a heterogeneous population for a wide variety of activities beyond walking. The hip exoskeleton possesses the range of motion 130° flexion, 40° extension, 90° abduction, and 60° adduction, encompassing the maximum range of the human hip joints [26]. Therefore, our hip exoskeleton can assist walking, sitting, squatting, and stair climbing. For a human of 75 kg walking at 1.6 m/s, the peak torque and the speed of the hip joint are 97 Nm (1.3 Nm/Kg) and 2.3 rad/s, respectively [27]. [28, 29] show that a hip exoskeleton with 12 Nm torque resulted in 15.5% metabolics reduction for uphill walking. To gain higher torque output, our hip exoskeleton is required to provide 20 Nm torque (about 20% of the biological joint moment). The design requirements are summarized in Table. I.

TABLE I.  DESIGN PARAMETERS OF THE HIP EXOSKELETON

| Parameters | Walking | Desired | Actual |
|---|---|---|---|
| Hip flexion/extension (°) | 32.2/22.5 | 32.2/22.5 | 130/40 |
| Hip abduction/adduction (°) | 7.9 /6.4 | 7.9 /6.4 | 90/60 |
| Max hip joint moment (Nm) | 97 | 19.4 | 20 |
| Max hip joint speed (rad/s) | 2.3 | 3 | 19.6 |
| Exoskeleton weight (kg) | —— | 5 | 3.4 |

## III. QUASI-DIRECT DRIVE ACTUATION

Quasi-direct drive actuation [10, 30] is a new paradigm of robot actuation design that leverages high torque density motors with a low ratio transmission mechanism. It is recently studied for legged robots [10], and wearable robots [31, 32] and have the benefits including a simplified mechanical structure, reduced mechanical inertia, and embodied physical intelligence (high backdrivability and high bandwidth without relying on active control) to improve safety and robustness in unstructured environments.

Due to the low gear ratio transmission, the primary limitation of the quasi-direct drive actuation paradigm is the output torque density. To overcome this barrier, a custom quasi-direct drive actuator is developed that provides high nominal torque, well-integrated, reduced speed (than conventional high-speed motors but fast enough for human movement assistance), and high torque-inertia ratio.

### A. High Torque Density Motor

In terms of topology, [33] and [34] show that the exterior rotor and concentrated winding type brushless DC (BLDC) electric motor has higher torque density. Besides topology design, we perform both mechanical and electromagnetic optimization to further improve its performance. As shown in Fig. 2(a), the custom motor uses fractional-slot type winding, which enables to reduce the cogging torque to $25 \pm 12.5$ mNm. Compared to commercially available motors (Maxon EC90, 12 pole pairs [17]) and research prototypes (10 pole pairs and 18 rotor slots [31]), our design has 21 pole pairs and 36 rotor slots which have higher output torque capability and lower cogging torque. The copper loss is 32.6 W, which is 26% less than the research prototypes (44.3 W) [31].

Our motor winding is attached to the stators and the rotor only consists of 1 mm thick permanent magnet chips and the rotor cover. The lightweight exterior rotor reduces rotary inertia and increases the torque-inertia ratio. Moreover, the customized motor uses sintered Neodymium Iron Boron (NdFeB) permanent magnets, which can reach 1.9 Tesla magnetic field intensity, as shown in Fig. 2 (b).

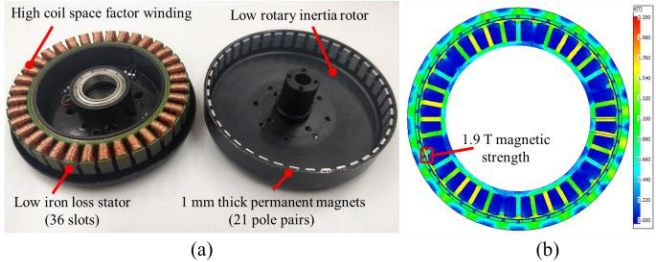

Fig. 2. (a) The custom-designed pancake exterior rotor type and concentrated winding brushless DC electric motor (BLDC) motor. The fractional slot design (36 slots, 21 pole pairs) reduces cogging torque and minimizes copper loss [31]. (b) By using sintered Neodymium Iron Boron permanent magnets, the finite element analysis simulation result shows under no current in the winding condition, the magnetic strength of the stator can reach 1.9 Tesla.

The customized high torque density motor weighs 274 g with 0.9 kg·cm$^2$ rotor inertia and can generate about 2.165 Nm nominal torque. This results in 8 Nm/kg nominal motor torque density and 2.2 Nm/ kg·cm$^2$ torque-inertia ratio. Compared with the benchmark motor (Maxon EC flat 505592, 0.7 Nm/kg nominal torque density, 0.14 Nm/ kg·cm$^2$ torque-inertia ratio) [17], our customized motor has about 11 times higher torque density and about 16 times higher torque-inertia ratio. The detailed specifications of the motor can be found in Table II and Table III.

### B. Quasi-direct Drive Actuator

To accommodate the dimension and mass requirement of exoskeleton design, a fully-integrated actuator can reduce mass, dimension, and ensure the operation safety. This is crucial to overcome the barriers for widespread adoption of wearable robots [35, 36] that have been typically considered heavy and bulky. We develop a compact direct-drive actuator with fully-integrated components, as shown in Fig. 3 using the custom-designed motor.

The overall actuator weight is 777 g, and it includes a high torque density motor, an 8:1 ratio planetary gear, a 14 bits high accuracy magnetic encoder, and a 10-60 V wide range input motor driver and controller. Moreover, the low-level control loop is implemented in the driver-control electronics to realize low-level control, including position, velocity, and current control. High-level control devices can send a command to read and write the real-time information actuator through the Controller Area Network (CAN bus) communication protocol. With a nominal voltage of 42 V, the actuator reaches a nominal speed of 188 RPM (19.7 rad/s). Thanks to the quasi-direct drive design using low gear ratio transmission, the actuator has low output inertia (57.6 kg·cm$^2$), which is essential to render low impedance to minimize the resistance to natural human movements.

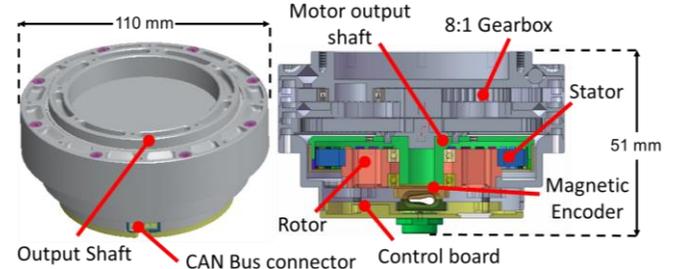

Fig. 3. The fully-integrated quasi-direct drive actuator consists of a high torque density motor, an 8:1 gearbox, a magnetic encoder, and control electronics. The motor is designed to be compact ($\Phi$110 mm × 52 mm height), lightweight (777 g), and able to generate high torque (17.5 Nm nominal torque and 42 Nm peak torque).

## IV. DESIGN AND MODELING OF QUASI-DIRECT DRIVE EXOSKELETON

This section presents the mechanical and electrical design of the exoskeleton, modeling of human-robot interaction, and analysis of mechanical versatility. Our quasi-direct drive actuation robot is benchmarked with conventional actuation and SEAs to realize minimal resistance to human movements and high dynamic performance to understand human-robot interaction.

### A. High Backdrivability and High-Bandwidth Exoskeleton

Since sagittal plane movements produce the dominant positive power of hip joints during walking and squatting, the robot is designed to have one active degree of freedom (DOF) and passive DOFs in frontal and transverse planes to ensure a natural range of motion. The hinge joints that connect the motor housings and the waist frame enable passive DOF in the frontal plane (i.e., abduction and adduction). The elastic thigh straps enable passive DOF in the transverse plane (i.e., internal/external rotations). The mechanical system of the hip exoskeleton is mainly composed of a waist frame, two actuators, two torque sensors, and two thigh braces, as shown in Fig. 1. The robot is symmetric about the sagittal plane. We design a waist frame to anchor the two actuators. The curvature of the waist frame is conformal to the wearer's pelvis, enabling uniform force distribution on the human body. A wide waist belt is chosen to attach the waist frame to the user, aiming to maximize the contact area and reduce pressure on the human. The two motors are fixed to the waist frame by the motor housings. The two actuators work in the sagittal plane to assist the flexion and extension of the hip joints. A custom torque sensor (± 40 Nm full scale and ± 0.1 Nm resolution) is assembled to the output flange of the actuator to measure output torque, as shown in Fig.1. The thigh brace is connected to the medial side of the torque sensor and transmits the actuator torque to the wearer's thigh. This curved structure of the thigh brace enables the assistive force on the wearer's thigh to be perpendicular to the frontal plane on the centerline of the thigh. The wearer does not suffer from the shear forces, which typically induce discomfort. Since hip extension moment is slightly larger than flexion moment during walking, the metallic thigh brace is used to generate extension assistance (Fig. 1 and 4). To reduce the loss of moment transmission while ensuring a comfortable human-exoskeleton contact, we used non-extensible fabric straps with foam for flexion assistance. Our hip exoskeleton supports the wearer to perform normal activities that require a large range of motion, including stair climbing, sit-to-stand, and squatting, as shown in Fig. 4.

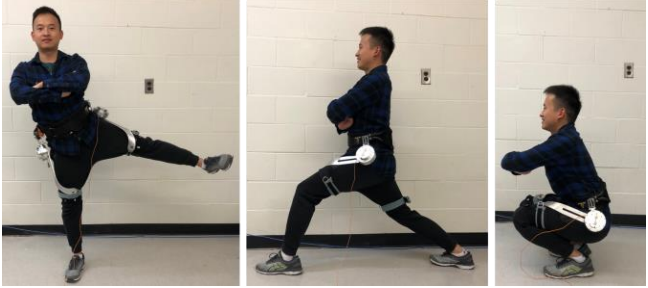

Fig. 4. Demonstration of the range of motion of the robot. The hip exoskeleton is designed with large ranges of motion to support the wearer to perform various activities.

The electrical system of the hip exoskeleton supports high-level torque control, motor control, sensor signal conditioning, data communication, and power management. The local motor controller is developed based on a motor driver and a DSP microcontroller to measure the motor motion status and realize motor current, velocity, and position control. The high-level microcontroller runs on Arduino Due to torque control. It acquires a lower-limb attitude from the wireless IMU sensors and conditions torque signals from the custom loadcells in real-time.

*B. Human-Exoskeleton Coupled Dynamic Model*

In this section, quasi-direct drive (QDD) and the human-exoskeleton interaction is modeled in comparison with series elastic actuation (SEA) to understand the interaction dynamics by characterizing the mechanical versatility of the robot in terms of backdrivability, control bandwidth, and torque capability as shown in Fig. 5. The system can be approximated as a mechanical system consisting of a motor, a gearbox, transmission, and wearable structures attached to the human hip joint. The connections between these modules are modeled as springs and dampers to represent the force and motion transmission.

The motor generates the torque $\tau_m$ with an input voltage $V$. The dynamics of the motor electrical system can be characterized by the winding resistance $R$ and inductance $L$. The motor torque $\tau_m$ is generated by current $i$ and the torque constant of motor is $k_t$. The back-electromagnetic force $V_b$ is proportional to the motor velocity $\dot{\theta}_m$ and the constant of back-electromagnetic force is $k_b$. The governing equation of the motor electrical system is

$$V - V_b = L\frac{di}{dt} + Ri \qquad (1)$$
$$\tau_m = k_t i \qquad (2)$$
$$V_b = k_b \dot{\theta}_m \qquad (3)$$

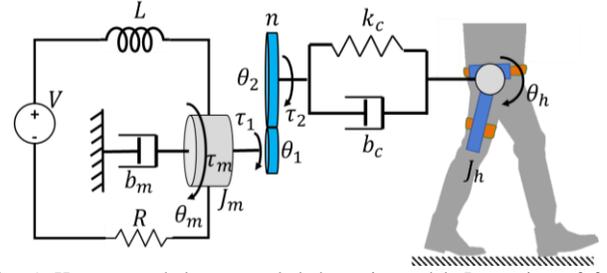

Fig. 5. Human-exoskeleton coupled dynamic model. It consists of four modules: motor system, speed reduction mechanism, transmission, wearable structure, and human leg.. $V$: motor nominal voltage; $R$: motor resistance (phase to phase); $L$: motor inductance (phase to phase); $b_m$: motor friction coefficient; $\tau_m$: torque developed by the motor; $\theta_m$: motor rotation angle; $J_m$: motor inertia; $\tau_1$: motor output torque; $\theta_1$: motor rotation angle before the gear reduction; $n$: gear ratio; $\tau_2$: gearbox output torque; $\theta_2$: gearbox output rotation angle; $k_c$: transmission stiffness; $b_c$: transmission damping; $J_h$: human leg inertia.

Meanwhile, the damping coefficient of viscous friction from the rotor is denoted by $b_m$. The torque applied to the input shaft before the gear reduction is $\tau_1$. The motor governing equation:

$$\tau_m = J_m \ddot{\theta}_m + b_m \dot{\theta}_m + \tau_1 \qquad (4)$$

where $J_m$ denotes the moment of inertia of the motor rotor around its rotation axis and $\theta_m$ denotes the motor angle.

The motor outputs the torque $\tau_1$ to the input shaft of the gearbox with a gear ratio $n$: 1. The output angle is reduced, and the torque is amplified through the gearbox.

$$\theta_1 = \theta_m, \ \theta_2 = \frac{\theta_1}{n}, \ \tau_2 = n\tau_1 \qquad (5)$$

In eq. (5), $\theta_1$ and $\theta_2$ denote the rotation angle of the input shaft and output shaft in the gearbox respectively, and $\tau_1$ and $\tau_2$ denote the torque applied on the input shaft and output shaft, respectively.

The transmission of the exoskeleton can be designed with rigid linkages, spring, cable-pulley systems, cable-textile systems which can be modeled as a transmission stiffness and damping, denoted by $k_c$ and $b_c$ respectively.

$$\tau_2 = b_c s(\dot{\theta}_2 - \dot{\theta}_h) + k_c(\theta_2 - \theta_h) \qquad (6)$$

The human thigh is governed by the equation

$$J_t \ddot{\theta}_h + b_c s(\dot{\theta}_h - \dot{\theta}_2) + k_c(\theta_h - \theta_2) = \tau_l \qquad (7)$$

where $J_t$ is the inertia of the thigh orthosis and human limb. $\theta_h$ is the hip rotation angle. $\tau_l$ is the human torque generated by the muscles. $\tau_a$ is the torque applied on the human thigh can be calculated as equation (8) and $\tau_a$ can be assistive torque or resistive torque.

$$\tau_a = \tau_2 = b_c s\left(\frac{\dot{\theta}_m}{n} - \dot{\theta}_h\right) + k_c\left(\frac{\theta_m}{n} - \theta_h\right) \qquad (8)$$

Let $s$ be the Laplace variable, and the system block diagram can be modeled as Fig. 6.

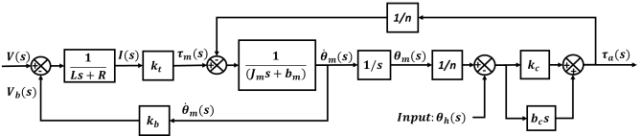

Fig. 6. The block diagram of human-exoskeleton open loop system

Assuming the initial condition $\theta_h(0), \dot{\theta}_h(0),$ and $V(0)$ is equal to zero and neglect the inductance $L$ due to its small value [37]. The Laplace transform of assistive torque $T_a(s)$ is related to the Laplace transform of the hip rotation angle $\theta_h(s)$ and input voltage $V(s)$, shown as equation (9).

$$\tau_a(s) = n_1(s)\left[\frac{n_2(s)}{d(s)}V(s) + \frac{n_3(s)}{d(s)}\theta_h(s)\right] \qquad (9)$$
$$= G_1(s)V(s) + G_2(s)\theta_h(s)$$
$$d(s) = J_e s^2 + b_e s + k_e$$
$$J_e = n^2 J_m R, b_e = n^2 R b_m + n^2 k_b k_t + R b_c, k_e = R k_c$$
$$n_1(s) = (b_c s + k_c); n_2(s) = n k_t$$
$$n_3(s) = -n^2[J_m R s^2 + (R b_m + k_b k_t)s]$$

$G_1(s)$ is the transfer function that defines the relationship between the voltage input $V(s)$ and human-robot interaction torque $\tau_a(s)$.

$$G_1(s)|_{\theta_h(s)=0} = \frac{\tau_a(s)}{V(s)} = \frac{n_1(s)n_2(s)}{d(s)} \qquad (10)$$

$G_2(s)$ is the transfer function that defines the relationship between the human motion input $\theta_h(s)$ and human-robot interaction torque $\tau_a(s)$.

$$G_1(s)G_2(s)|_{V(s)=0} = \frac{\tau_a(s)}{\theta_h(s)} = \frac{n_1(s)n_3(s)}{d(s)} \qquad (11)$$

The natural frequency $\omega_n$ of the open-loop torque control for the second-order system is

$$\omega_n = \sqrt{\frac{k_e}{J_e}} = \sqrt{\frac{R k_c}{n^2 J_m R}} = \sqrt{\frac{k_c}{n^2 J_m}} \qquad (12)$$

The effective moment of inertia $J_e$ is equal to $n^2 J_m$ and the natural frequency $\omega_n$ of the open-loop torque control is directly modulated by gear ratio $n$ and the transmission stiffness $k_c$ and moment of inertia of the motor $J_m$. In [12], it also concluded that the reflected actuator inertia directly modulates the effective mass, and small reflective actuator inertia will increase the open-loop force control bandwidth. The smaller gear ratio, the smaller movement of inertia of the motor, and the larger transmission stiffness will generate higher natural frequency.

To investigate the backdrivability, $V(s)$ is set to zero to identify the property of the passive mechanism, and the output resistive torque and output link impedance are used to analyze the back drivability. The resistive torque $\tau_a$ induced by the human motion $\theta_h(s)$ can be derived through equation (13). The output link impedance is derived as equation (14).

$$\tau_a(s) = G_2(s)\theta_h(s) =$$
$$\frac{-(b_c s + k_c)n^2[J_m R s^2 + (R b_m + k_b k_t)s]}{n^2[J_m R s^2 + (R b_m + k_b k_t)s] + R b_c s + R k_c}\theta_h(s) \qquad (13)$$

$$Z_o(s) = \frac{\tau_a(s)}{\dot{\theta}_h(s)} = \frac{\tau_a(s)}{s\theta_h(s)} \qquad (14)$$

As the gear ratio is sufficiently small, the resistive torque can be approximated as zero, shown as equation (15).

$$\lim_{n \to 0} \tau_a(s) = 0 \qquad (15)$$

As the gear ratio is large, the resistive torque will be approximated to equation (16) and the transmission damping constant $b_c$ and stiffness constant $k_c$ are the dominated terms.

$$\lim_{n \to \infty} \tau_a(s) \approx -(b_c s + k_c)\theta_h(s) \qquad (16)$$

When the gear ratio is 1, the resistive torque is shown in equation (17), and we can observe the resistive torque are modulated by the gear ratio $n$, the transmission damping constant $b_c$, the transmission stiffness constant $k_c$, the motor inertia $J_m$, the motor damping constant $b_m$, the motor resistance $R$, the motor torque constant $k_t$, and the back EMF constant $k_b$.

$$\tau_a(s)|_{n=1}$$
$$= -\frac{(b_c s + k_c)(J_m R s^2 + R b_m s + k_b k_t s)}{J_m R s^2 + (R b_m + k_b k_t + R b_c)s + R k_c}\theta_h(s) \qquad (17)$$

Through the equations (15), (16), and (17), a highly backdrivable mechanism, low output link impedance, and low resistive torque can be achieved by a smaller gear ratio $n$, a smaller damping constant $b_c$, or a smaller stiffness constant $k_c$.

## C. Actuation Paradigm Based Hip Exoskeleton Benchmark

Based on the human-robot interaction model, we benchmark the performance of state-of-the-art hip exoskeletons in terms of actuation paradigms, including conventional actuation, SEA, and QDD. The difference between conventional actuation and SEA is the transmission stiffness. The difference between the QDD and conventional actuation is the gear ratio and the motor type. The representative devices include 1) the conventional actuation based hip exoskeleton [13]; 2) the SEA based hip exoskeleton [17]; 3) our quasi-direct drive hip exoskeleton. Table II shows the key parameters of the actuators from these three representative hip exoskeletons. Note that the parameters that affect the backdrivability vary within a specific range [10, 38]. For simplicity of benchmark comparison, we set the transmission damping constant (0.01 Nm·s/rad) in the simulation model and use a large transmission stiffness (i.e., 500 Nm/rad) for the conventional and QDD actuators, and set a relatively small transmission stiffness (i.e., 120 Nm/rad) for the series elastic actuator due to its spring component.

To benchmark the versatility of three actuation methods, we use resistive torque (backdrivability) and bandwidth of torque control as metrics to demonstrate the benefit of low gear ratio and high torque density motor and the results are as shown in

Fig. 7. To approximate the real condition, the Simulink model used the saturation function to limit the maximum voltage. The maximum voltage of conventional actuation, SEA, and QDD is set as 24V, 48V, and 42V, respectively.

TABLE II. PARAMETERS IN HUMAN-EXOSKELETON COUPLED MODEL

| Parameter | Unit | Conventional | SEA | QDD |
|---|---|---|---|---|
| Motor | - | EC 45 | EC 90 | Ours |
| Nominal voltage | V | 24 | 48 | 42 |
| Nominal current | A | 3.21 | 2.27 | 7.5 |
| Nominal Torque | Nm | 0.128 | 0.533 | 2.165 |
| Motor resistance | Ω | 0.608 | 2.28 | 0.58 |
| Motor inductance | mH | 0.46 | 2.5 | 0.21 |
| Motor friction coefficient | Nm·s/rad | 0.01 | 0.01 | 0.08 |
| Torque constant | Nm/A | 0.0369 | 0.217 | 0.2886 |
| Motor inertia | g·cm$^2$ | 181 | 3060 | 895 |
| Gear ratio | - | 50:1 | 100:1 | 8:1 |
| Transmission stiffness | Nm /rad | 500 | 120 | 500 |
| Transmission damping | Nm·s/rad | 0.01 | 0.01 | 0.01 |

For the bandwidth analysis of the closed-loop torque control, the block diagram is shown in Fig. 6, where the input reference torque is a chirp signal (0~100 Hz) with 5 Nm magnitude, and the input human hip angle is set as zero (simulation for the fixed output link).

For the resistive torque simulation (i.e., backdrivability), the block diagram is shown in Fig. 7 and the input human hip angle is set as a chirp signal (0~1 Hz) with 10-degree magnitude, and the voltage source V(s) is set as zero. The simulation results (Fig. 8) demonstrate that the designed quasi-direct drive actuation with lower gear ratio and high torque density motor achieves mechanical versatility in terms of high bandwidth (73.3 Hz) and high backdrivability (backdrive torque 0.97 Nm). Both the conventional actuation and SEA have compromised bandwidth or backdrivability.

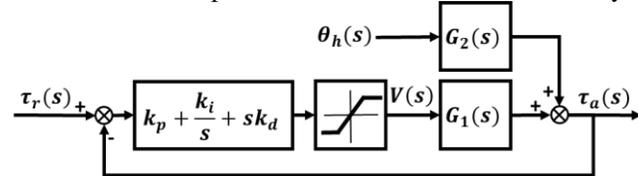

Fig. 7. The block diagram of closed-loop torque control in the human-exoskeleton system.

In addition to those two metrics, a high-performance portable hip exoskeleton also requires high torque output and low mass. In Table III, we listed the comparison results of these three actuators' performance. The key factors in evaluating the actuator performance are torque inertia ratio, bandwidth, backdrivability, and output torque density. It shows that our quasi-direct drive actuator not only outperforms in terms of bandwidth and backdrivability, it also possesses a significantly higher torque inertia ratio without sacrificing torque density as its two counterparts (high gear ratio can easily enhance torque).

Due to the simplification of the model and variability of parameters (e.g., human limb inertia), the simulation results may deviate from the actual values. Though the SEAs model presented in this simulation may not represent novel methods of SEAs [18, 19] (e.g., we can use those novel methods of SEAs for QDD actuation too), our analysis is able to capture the underlying characteristics of the three actuation paradigms.

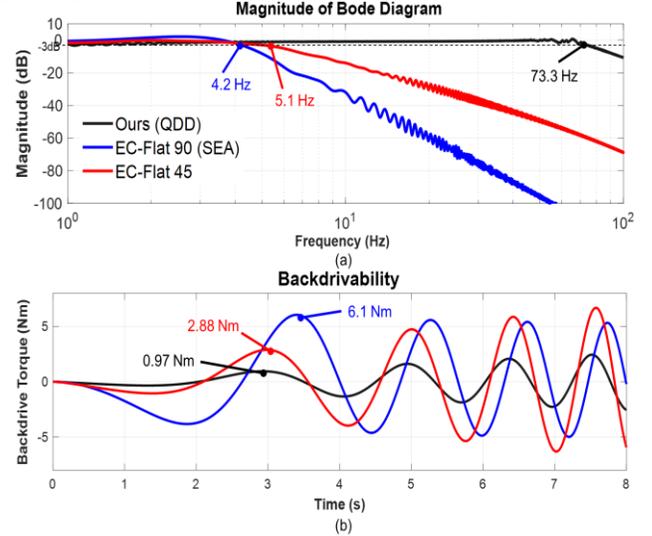

Fig. 8. The bandwidth and backdrivability simulation result based on the human-robot interaction model. The result shows that our quasi-direct drive actuator has a theoretical 73.3 Hz bandwidth and 0.97 Nm backdrivability. Compared with the other two actuation paradigms, the quasi-direct drive actuation demonstrates significant improvements in terms of bandwidth and backdrivability.

TABLE III. COMPARISON OF HIP EXOSKELETONS PERFORMANCE OF THREE ACTUATOIN PARADIGMS

| Parameter | Unit | Conventional [13] | SEA [17] | QDD (this work) |
|---|---|---|---|---|
| Output nominal torque | Nm | 8 | 40 | 17.5 |
| Actuator mass | kg | ~0.50 | 1.80 | 0.77 |
| Actuator nominal torque density | Nm/kg | 16 | 22.2 | 20.7 |
| Control bandwidth | Hz | 5.1 | 4.2 | 73.3 |
| Backdrive torque | Nm | 2.88 | 6.10 | 0.97 |

## V. CONTROL STRATEGIES

The control system is based on a hierarchical architecture composed of a high-level control layer that robustly detects gait intention to compensate for the uncertainty due to varying walking and squatting speed (Fig. 9), a middle-level control that generates assistance torque profile (Fig. 9), and a low-level control layer that implements a current-based torque control.

To compensate for the uncertainties caused by changing gait speeds, an algorithm recently developed in our lab that uses a data-driven method [39] with a neural network regressor to estimate the walking and squatting percentage in real-time by the signals from two inertial measurement units (IMUs). These two IMUs are mounted on anterior of both left and right thighs (Fig. 1). These sensors provide the motion information, including Euler angle, angular velocity, and acceleration at a frequency of 200 Hz. Each signal channel is scaled to have zero mean and unit variance. The motion information during the last 0.4 s sliding time window constitutes the input vector of the neural network for both offline training process and online control. The neural network used in this algorithm has one hidden layer with 30 neurons as well as a sigmoid activation function and deploys the Xavier initialization of network weights. We collect the walking and squatting data from three able-bodied subjects at several different speeds, and our algorithm could achieve an $R^2 = 0.997$ on the test set.

After obtaining the gait percentage, the middle-level controller calculates the assistive torque according to a predefined torque profile expressed as a look-up table. The predefined torque profile for walking is generated by the human biological model in [40], and squatting is a sine wave. Searching the gait percentage in the look-up table and applying an interpolation calculation could produce the desired assistance torque.

The low-level torque control architecture is composed of an inner-loop control and an outer-loop control. The inner loop implements motor current control in the local motor controller. The outer loop implements the torque control in Arduino Due whose feedback signal is from motors, loadcells, and IMU-based gait detection. The simplicity of the control algorithm corroborates the underlying philosophy about "design for control", namely robots can be designed to simplify the control while ensuring high performance.

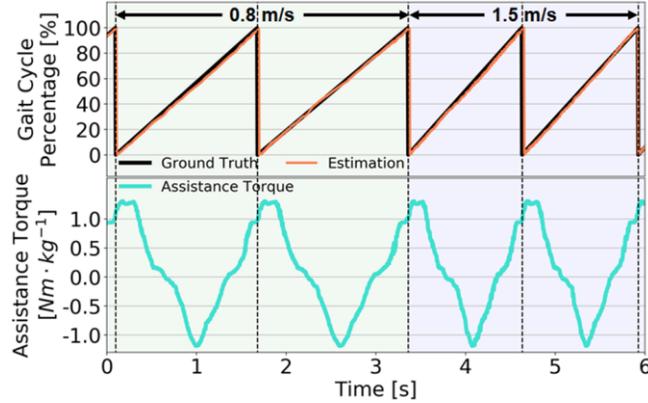

Fig. 9. (Top) The orange line represents the estimated gait cycle percentage by our regression method, while the black line represents the ground truth calculated offline by insole signals. The robust gait estimation is able to compensate for the disturbance due to walking speed changes. The estimated gait matches well with the ground truth ($R^2 = 0.997$). (Bottom) The turquoise line shows 100% of the biological torque generated by our algorithm.

## VI. EXPERIMENTAL RESULTS

To demonstrate the backdrivability and bandwidth of the hip exoskeleton, we conducted five experiments on the actuator and the hip exoskeleton to systematically characterize the mechanical versatility of the device.

### A. Motor Nominal Current Evaluation

The actuator output capability is highly limited by the motor's winding temperature. To evaluate the actuator working current performance, we operated the actuator continuously in the stall mode under different output currents. The stator temperature was measured by an embedded temperature sensor, and the actuator surface temperature was measured by a portable FLIR® thermal camera. The experiment was performed in a 22 °C lab environment without external heat dissipation. The maximum operating time was set to 15 mins and the maximum temperature of the stator to 100 °C. Fig. 10(a) shows the stator temperature changes with time under different current conditions. Fig. 10(b) shows the thermal camera image for 15 mins when the actuator stalls under the nominal current and the highest temperature of the actuator shell is 62.7 °C. The experiment demonstrates that the custom actuator is able to produce a nominal torque of 17.5 Nm under 7.5 A nominal current and about 42 Nm peak torque under 18 A current based on the calibrated torque constant.

### B. Exoskeleton Bandwidth Evaluation

The bandwidth experiment of the torque control was carried out to obtain the Bode plot. A chirp signal was used as the reference torque whose magnitudes are set as 10 Nm, 15 Nm, and 20 Nm. The Bode plot shows the bandwidth was 57.8 Hz, 59.3 Hz, and 62.4 Hz, respectively, as shown in Fig. 11. Compared with the model-based control bandwidth (73.3 Hz) simulation results in Fig. 8, the bandwidth obtained through experimental results is relatively low (62.4 Hz). This is primarily due to lower damping and to the effect of the actuator backlash in the hardware prototype. The control bandwidth is much higher than the requirement of human walking, but this is useful for agile human activities, e.g., running and balance control to unexpected external disturbance. Compared with the exoskeletons using SEA [16] with 5 Hz bandwidth, the high control bandwidth robot is safer and more robust to uncertainties.

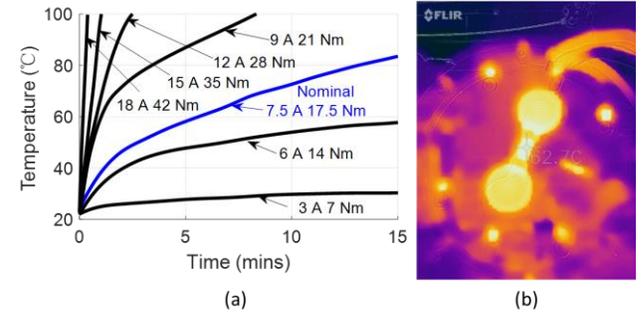

Fig. 10. (a) Stator temperature over time for different current conditions; (b) Thermal image after 15 min of continuous 7.5 A current operation. The actuator surface reached the maximum temperature of 62.7 °C.

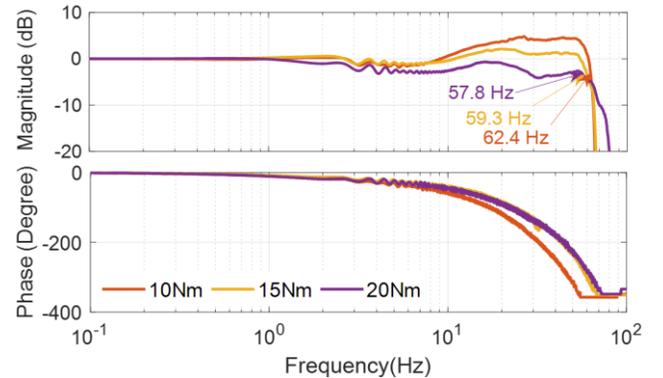

Fig. 11. Bode plot for 10 Nm, 15 Nm, and 20 Nm torque control. The high bandwidth (highest value is 62.4 Hz) demonstrates the ability to handle high-frequency movements of humans in comparison with state-of-the-art results (for instance, a 5 Hz bandwidth is reported in [16]).

### C. Exoskeleton Backdrive Torque Evaluation

To demonstrate the backdrivability of the exoskeleton, we tested the backdrive torque in unpowered mode. One able-bodied subject wore the exoskeleton, and the hip joint moved with a 32.2° range of motion at 1 Hz while the actuator off. The measured rotation angle and the backdrive torque are presented in Fig. 12. The hip exoskeleton has a very low backdrive torque (~0.4 Nm), which shows better backdrivability than SEA and QDD based exoskeletons[16, 31].

*D. Torque Tracking for Walking and Squatting Assistance*

An able-bodied subject (26-year-old male, 70 kg) wore the hip exoskeleton and walked on a treadmill with speed varying from 0.8 to 1.4 m/s and squatted on the ground with 2 s cadence. The predefined amplitude of the torque density is 1.3 Nm/kg, resulting in a total of 91 Nm. The control test with 22% of biological hip joint movement during walking and squatting was performed to investigate the tracking performance. The torque of the proposed device (about 20 Nm) is higher than the hip device (12 Nm torque) [13].

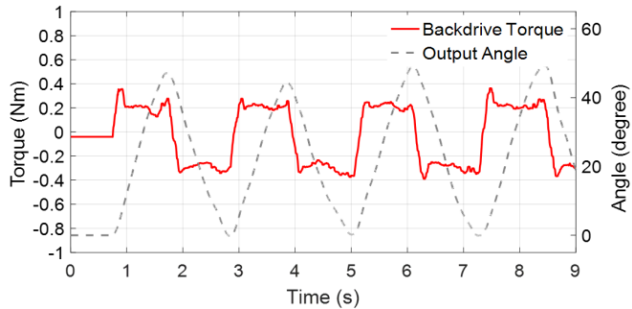

Fig. 12. The backdrivability of the hip exoskeleton in the unpowered mode. The maximum torque of the mechanical resistance is approximately 0.4 Nm in comparison with state-of-the-art results in [16]with 2 Nm and [31]with about 1 Nm.

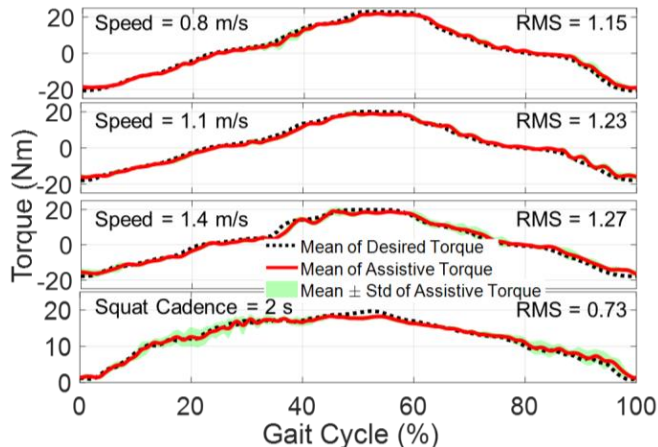

Fig. 13. Torque tracking performance of ±20 Nm assistance during walking and squatting tests. The mean of actual assistive torque (red) tracked the desired torque (black dash) well. The RMS errors of torque tracking (0.8 m/s, 1.1 m/s and 1.4 m/s walking, 2 s cadence squatting) were 1.15 Nm, 1.23 Nm, 1.27 Nm and 0.73 Nm respectively (5.75%, 6.15%, 6.35% and 3.65% of the peak torque). The tracking accuracy is superior to the SEA based exoskeleton (±10 Nm and 21.5 % error) [16].

A total of 15 tests for the same torque profile for each of the walking and squatting motions were performed. Torque profile, according to the gait cycle, was generated from the high-level controller, and the tracking performance of the hip assistance is shown in Fig. 13. The average RMS error between the desired and actual torque trajectory in 60 tests is 1.09 Nm (5.4% error of the maximum desired torque). The results indicate that the torque controller can follow the torque profile with high accuracy to assist human walking and squatting.

## VII. DISCUSSION AND CONCLUSION

This paper presents a quasi-direct drive actuation that is composed of custom high torque density motor and low ratio gear transmission (8:1). It can generate high torque (17.5 Nm nominal torque, 42 Nm peak torque) and weighs 777g. A lightweight bilateral portable hip exoskeleton is developed based on this actuation paradigm, which exhibits mechanical versatility through its high backdrivability (unpowered mode 0.4 Nm backdrive torque) and high bandwidth (62.4 Hz) without using complicated control methods. Both benchmark simulation and experimental study indicate that our exoskeleton with this novel actuation method outperforms state-of-the-art performance [16-20] in terms of device mass, bandwidth, backdrivability, and torque tracking performance.

The limitation of this work is that there is no human performance study as it focuses on the quantitative characterization of the quasi-direct drive actuation for exoskeletons. We will study human performance (e.g., kinetics and electromyography) in our future work. We believe that our enabling technologies and modeling framework offer essential tools to understand human-robot interaction and push the limits to improve locomotion economy [5] as personal mobility assistance devices.